\documentclass[runningheads]{llncs}
\usepackage{graphicx}
\usepackage{amsmath}
\usepackage{amssymb}
\usepackage{xcolor}
\usepackage{algorithm}
\usepackage{algorithmic}
\usepackage[normalem]{ulem}
\useunder{\uline}{\ul}{}

\begin{document}
\title{State Representation Learning for Goal-Conditioned Reinforcement Learning}
%
%
\author{Lorenzo Steccanella\inst{1} \and Anders Jonsson\inst{1}}
\authorrunning{Steccanella et al.}
%
\institute{Dept.~Information and Communication Technologies, Universitat Pompeu Fabra, Barcelona, Spain
\email{\{lorenzo.steccanella, anders.jonsson\}@upf.edu}}
\maketitle              

\begin{abstract}

This paper presents a novel state representation for reward-free Markov decision processes. The idea is to learn, in a self-supervised manner, an embedding space where distances between pairs of embedded states correspond to the minimum number of actions needed to transition between them. Compared to previous methods, our approach does not require any domain knowledge, learning from offline and unlabeled data.
We show how this representation can be leveraged to learn goal-conditioned policies, providing a notion of similarity between states and goals and a useful heuristic distance to guide planning and reinforcement learning algorithms.
Finally, we empirically validate our method in classic control domains and multi-goal environments, demonstrating that our method can successfully learn representations in large and/or continuous domains.

\end{abstract}

\section{Introduction}

In reinforcement learning, an agent attempts to learn useful behaviors through interaction with an unknown environment. By observing the outcome of actions, the agent has to learn from experience which action to select in each state in order to maximize the expected cumulative reward.

In many applications of reinforcement learning, it is useful to define a metric that measures the similarity of two states in the environment. Such a metric can be used, e.g., to define equivalence classes of states in order to accelerate learning, or to perform transfer learning in case the domain changes according to some parameters but retains part of the structure of the original domain. A metric can also be used as a heuristic in goal-conditioned reinforcement learning, in which the learning agent has to achieve different goals in the same environment. A goal-conditioned policy for action selection has to reason not only about the current state, but also on a known goal state that the agent should reach as quickly as possible.

\begin{figure}[t]
    \centering
    \includegraphics[width=8cm]{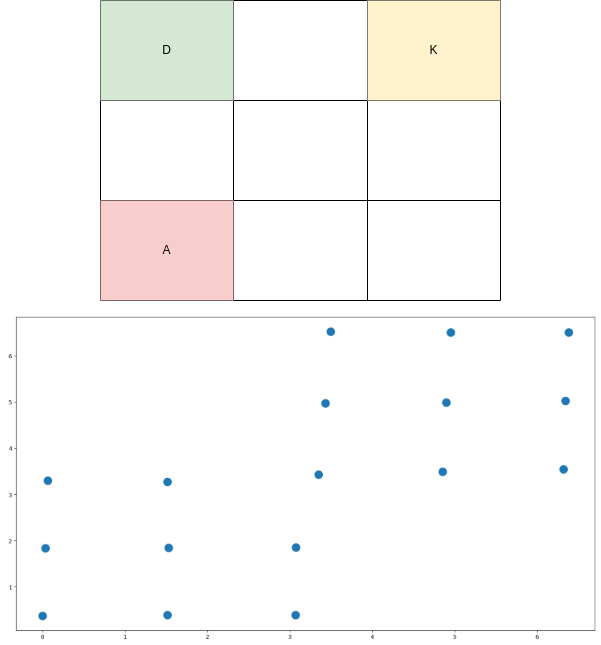}
    \caption{Top: a simple gridworld where an agent has to pick up a key and open a door (key and door positions are fixed). Bottom: the learned state embedding $\phi$ on $\mathbb{R}^2$. The state is composed of the agent location and whether or not it holds the key.}
    \label{fig:KeyDoorRepr}
\end{figure}

In this work, we propose a novel algorithm for computing a metric that estimates the minimum distance between pairs of states in reinforcement learning. The idea is to compute an embedding of each state into a Euclidean space (see Fig.~\ref{fig:KeyDoorRepr}), and define a distance between pairs of states equivalent to the norm of their difference in the embedded space. We formulate the problem of computing the embedding as a constrained optimization problem, and relax the constraints by transforming them into a penalty term of the objective. An embedding that minimizes the objective can then be estimated via gradient descent.

The proposed metric can be used as a basis for goal-conditioned reinforcement learning, and has an advantage over other approaches such as generalized value functions. The domain of a generalized value function includes the goal state in addition to the current state, which intuitively increases the complexity of learning and hence the effort necessary to properly estimate a goal-conditioned policy. In contrast, the domain of the proposed embedding is just the state itself, and the distance metric is estimated by comparing pairs of embedded states.

In addition to the novel distance metric, we also propose a model-based approach to reinforcement learning in which we learn a transition model of actions directly in the embedded space. By estimating how the embedding will change after taking a certain action, we can predict whether a given action will take the agent closer to or further from a given target state. We show how to use the transition model to plan directly in embedded space. As an alternative, we also show how to use the proposed distance metric as a heuristic in the form of reward shaping when learning to reach a particular goal state.

The contributions of this work can be summarized as follows:

\begin{enumerate}
    \item We propose a self-supervised training scheme to learn a distance function by embedding the state space into a low-dimensional Euclidean space $\mathcal{R}^d$ where a chosen p-norm distance between embedded states approximates the minimum distance between the actual states.
    
    \item Once an embedding has been computed, we estimate a transition model of the actions directly in embedded space.
    
    \item We propose a planning method that uses the estimated transition model to select actions, and a potential-based reward shaping mechanism that uses the learned distance function to provide immediate reward to the agent in a reinforcement learning framework.
\end{enumerate}





\section{Related Work} \label{Related Work}

Our work relies on self-supervised learning to learn an embedding space useful for goal-conditioned reinforcement learning (GCRL). 


Goal-Conditioned Supervised Learning, or GCSL \cite{ghosh2020learning}, learns a goal-conditioned policy using supervised learning. The algorithm iteratively samples a goal from a given distribution, collects a suboptimal trajectory for reaching the goal, relabels the trajectory to add expert tuples to the dataset, and performs supervised learning on the dataset to update the policy via maximum likelihood.

Similar to our work, Dadashi et al.~\cite{DBLP:conf/icml/DadashiRVHPG21} learn embeddings and define a pseudo-metric between two states as the Euclidean distance between their embeddings. Unlike our work, an embedding is computed both for the state-action space and the state space. The embeddings are trained using loss functions inspired by bisimulation.

Tian et al.~\cite{tian2021mbold} also learn a predictive model and a distance function from a given dataset. However, unlike our work, the predictive model is learned for the original state space rather than the embedded space, and the distance function is in the form of a universal value function that takes the goal state as input in addition to the current state-action pair. Moreover, in their work they use ``negative'' goals assuming extra domain knowledge in the form of proprioceptive state information from the agent (e.g.~robot joint angles). Schaul et al.~\cite{schaul2015universal} also learn universal value functions by factoring them into two components $\phi:S\rightarrow\mathbb{R}$ and $\varphi:G\rightarrow\mathbb{R}$, where $G$ is the set of goal states. For a more comprehensive survey of goal-conditioned reinforcement learning, we refer to Liu et al.~\cite{https://doi.org/10.48550/arxiv.2201.08299}.

\section{Background} \label{Background}

In this section we introduce necessary background knowledge and notation.

\subsection{Markov Decision Processes}

A Markov decision process (MDP) \cite{puterman2014markov} is a tuple $\mathcal{M} = \langle S,A,P,r \rangle$, where $S$, $A$ denote the state space and action space, $P: S \times A \rightarrow \Delta(S)$ is a transition kernel and $r:S \times A \rightarrow \mathbb{R}$ is a reward function. At time $t$, the learning agent observes a state $s_t \in S$, takes an action $a_t \in A$, obtains a reward $r_t$ with expected value $\mathbb{E}[r_t] = r(s_t, a_t)$, and transitions to a new state $s_{t+1} \sim P(\cdot | s_t, a_t)$.

A stochastic policy $\pi:S\rightarrow\Delta(A)$ is a mapping from states to probability distributions over actions. The aim of reinforcement learning is to compute a policy $\pi$ that maximizes some notion of expected future reward.

In this work, we consider the discounted reward criterion, for which the expected future reward of a policy $\pi$ can be represented using a value function $V^\pi$, defined for each state $s\in S$ as
\[
V^\pi(s) = \mathbb{E} \left[ \left. \sum_{t=1}^{\infty} \gamma^{t-1} r(S_t,A_t) \right\vert S_1=s \right].
\]
Here, random variables $S_t$ and $A_t$ model the state and action at time $t$, respectively, and the expectation is over the action $A_t\sim\pi(\cdot|S_t)$ and next state $S_{t+1}\sim P(\cdot|S_t,A_t)$. The discount factor $\gamma\in(0,1]$ is used to control the relative importance of future rewards, and to ensure $V^\pi$ is bounded.

As an alternative to the value function $V^\pi$, one can instead model expected future reward using an action-value function $Q^\pi$, defined for each state-action pair $(s,a)\in S\times A$ as
\[
Q^\pi(s, a) = \mathbb{E} \left[ \left. \sum_{t=1}^{\infty} \gamma^{t-1} r(S_t,A_t) \right\vert S_1=s, A_1=a \right].
\]
The value function $V^\pi$ and action-value function $Q^\pi$ are related through the well-known Bellman equations:
\begin{align*}
V^\pi(s) &= \sum_{a\in A} \pi(a|s) Q^\pi(s,a),\\
Q^\pi(s,a) &= r(s,a) + \gamma \sum_{s'\in S} P(s'|s,a) V^\pi(s').
\end{align*}
The aim of learning is to find an optimal policy $\pi^*$ that maximizes the value in each state, i.e.~$\pi^*(s) = \arg\max_\pi V^\pi$. The optimal value function $V^*$ and action-value function $Q^*$ satisfy the Bellman optimality equations:
\begin{align*}
 V^{*}(s) &= \max _{a \in A} Q^*(s,a),\\
Q^{*}(s, a) &= r(s,a)+ \gamma \sum_{s^{\prime} \in S} P(s^{\prime} | s, a) V^*(s').
\end{align*}




\subsection{Goal-Conditioned Reinforcement Learning}

Standard RL only requires the agent to complete one task defined by the reward function. In  Goal-Conditioned Reinforcement Learning (GCRL) the observation is augmented with an additional goal that the agent is require to achieve when taking a decision in an episode \cite{schaul2015universal,andrychowicz2017hindsight}. 
GCRL augments the MDP tuple $\mathcal{M}$ with a set of goal states and a desired goal distribution $\mathcal{M}_G =\langle S,G,p_g,A,P,r \rangle$, where G is a subset of the state space $G\subseteq S$, $p_g$ is the goal distribution and the reward function $r: S \times A \times G \rightarrow \mathbb{R}$ is defined on goals $G$. Therefore the objective of GCRL is to reach goal states via a goal-conditioned policy $\pi: S \times G \rightarrow \Delta(A)$ that maximizes the expectation of the cumulative return over the goal distribution.

\subsubsection{Self-Imitation Learning}

When we consider the goal space to be equal to the state space $G=S$ we can treat any trajectory $t=\{s_0, a_0, ..., a_{n-1}, s_n\}$ and any sub-trajectory $t_{i,j} \in t$, as a successful trial for reaching their final states. Goal Conditioned Supervised Learning (GCSL) \cite{ghosh2020learning} iteratively performs behavioral cloning on sub-trajectories collected in a dataset $\mathcal{D}$ by learning a policy $\pi$ conditioned on both the goal and the horizon $h$:
\begin{equation*}
    J(\pi) = \mathbb{E}_\mathcal{D}[log \pi(a \mid s, g, h)].
\end{equation*}

\subsection{Reward Shaping}

An important challenge in reinforcement learning is solving domains with sparse rewards, i.e. when the immediate reward signal is almost always zero.


Reward Shaping attempts to solve this issue by augmenting a sparse reward signal $r$ with a reward shaping function $F$, $\overline{r} = r + F$. Based on this idea, Ng et al.~\cite{ng1999policy} proposed Potential-based reward shaping (PBRS) as an approach to guarantee policy invariance while reshaping the environment reward $r$. Formally PBRS defines $F$ as:

\begin{equation*}
    F = \gamma \Phi(s') - \Phi(s)
\end{equation*}
where $\Phi: S \rightarrow \mathcal{R}$ is a real-valued potential function.

\section{Contribution} \label{Contribution}

In this section we present our main contribution, a method for learning a state representation of an MDP that can be leveraged to learn goal-conditioned policies. We first introduce notation that will be used throughout, then present our method for learning an embedding, and finally show how to integrate the embedding in algorithms for planning and learning.



We first define the Minimum Action Distance (MAD) $d_{MAD}(s, s')$ as the minimum number of actions necessary to transition from state $s$ to state $s'$.

\begin{definition}
(Minimum Action Distance) Let $T\left(s' \mid \pi, s\right)$ be the random variable denoting the first time step in which state $s^{\prime}$ is reached in the MDP when starting from state $s$ and following policy $\pi$. Then $d_{MAD}(s, s')$ is defined as:
$$
d_{MAD}(s, s'):=\min _{\pi} min\left[T\left(s' \mid \pi, s\right)\right].
$$
\end{definition}


The Minimum Action Distance between states is a priori unknown, and is not directly observable in continuous and/or noisy state spaces where we cannot simply enumerate the states and keep statistics about the MAD metric. Instead, we will approximate an upper bound using the distances between states observed on trajectories. We introduce the notion of Trajectory Distance (TD) as follows:

\begin{definition}
(Trajectory Distance) Given any trajectory $t={s_0, ..., s_n} \sim \mathcal{M}$ collected in an MDP $\mathcal{M}$ and given any pair of states along the trajectory $(s_i, s_j) \in t$ such that $0 \leq i \leq j \leq n$, we define $d_{TD}(s_i, s_j \mid t)$ as
$$
d_{TD}(s_i, s_j \mid t) = (j - i),
$$
i.e.~the number of decision steps required to reach $s_j$ from $s_i$ on trajectory $t$.
\end{definition}

\subsection{State Representation Learning}

Our goal is to learn a parametric state embedding $\phi_\theta: S \rightarrow \mathcal{R}^d$ such that the distance $d$ between any pair of embedded states approximates the Minimum Action Distance from state $s$ to state $s'$ or vice versa.
\begin{equation}
d(\phi_\theta(s), \phi_\theta(s')) \approx min(d_{MAD}(s, s'), d_{MAD}(s', s)).
\end{equation}
We favour symmetric embeddings since it allows us to use norms as distance functions, e.g.~the L1 norm $d(z,y)=||z-y||_1$. Later we discuss possible ways to extend our work to asymmetric distance functions.

To learn the embedding $\phi_\theta$, we start by observing that given any state trajectory $t=\{s_0, ..., s_n\}$, choosing any pair of states $(s_i, s_j) \in t$ with $0 \leq i \leq j \leq n$, their distance along the trajectory represents an upper bound of the MAD.

\begin{equation} \label{eqn:inequalityMAD}
     d_{MAD}(s_i, s_j) \leq d_{TD}(s_i, s_j \mid t).
\end{equation}
Inequality \eqref{eqn:inequalityMAD} holds for any trajectory sampled by any policy and allows to estimate the state embedding $\phi_\theta$ offline from a dataset of collected trajectories $\mathcal{T} = \{t_1, ..., t_n\}$.
We formulate the problem of learning this embedding as a constrained optimization problem:

\begin{equation}
\begin{aligned}
\min_{\theta} \quad & \sum_{t\in\mathcal{T}}\sum_{(s,s')\in t} (\left\|\phi_\theta(s)-\phi_\theta(s')\right\|_{l} - d_{TD}(s, s' \mid t))^2,\\
\textrm{s.t.} \quad & \left\|\phi_\theta(s)-\phi_\theta(s')\right\|_{l} \leq d_{TD}(s, s' \mid t) \;\;\; \forall t \in \mathcal{T}, \forall (s,s')\in t.\\
\end{aligned}
\label{eqn:pf}
\end{equation}

Intuitively, the objective is to make the embedded distance between pairs of states as close as possible to the observed trajectory distance, while respecting the upper bound constraints. Without constrains, the objective is minimized when the embedding matches the expected Trajectory Distance $\mathbb{E}\left[d_{TD}\right]$ between all pairs of states observed on trajectories in the dataset $\mathcal{T}$. In contrast, constraining the solution to match the minimum TD with the upper-bound constrains $\left\|\phi_\theta(s)-\phi_\theta(s')\right\|_{l} \leq d_{TD}(s, s' \mid t)$ allows us to approximate the MAD. Evidently, the precision of this approximation depends to the quality of the given trajectories.


To make the constrained optimization problem tractable, we relax the hard constrains in \eqref{eqn:pf} and convert them into a penalty term in order to retrieve a simple unconstrained formulation that is solvable with gradient descent and fits within the optimization scheme of neural networks.

\begin{equation} \label{eqn:relaxed_pf}
\begin{aligned}
\min_{\theta} \quad & \sum_{t\in\mathcal{T}}\sum_{(s,s')\in t} \left[ \frac{1}{d_{TD}(s, s' \mid t)^2} (\left\|\phi_\theta(s)-\phi_\theta(s')\right\|_{l} - d_{TD}(s, s' \mid t))^2\right] + C,\\
\end{aligned}
\end{equation}
where $C$ is our penalty term defined as
\begin{equation*} \label{eqn:penalty_term}
\begin{aligned}
C = \sum_{t\in\mathcal{T}}\sum_{(s,s')\in t} \left[ \frac{1}{d_{TD}(s, s' \mid t)^2} \max \left(0, \left\|\phi_\theta(s)-\phi_\theta(s')\right\|_{l} - d_{TD}(s, s' \mid t) \right)^2\right].
\end{aligned}
\end{equation*}

The penalty term $C$ introduce a quadratic penalization of the objective for violating the upper-bound constraints $\left\|\phi_\theta(s)-\phi_\theta(s')\right\|_{l} <= d_{TD}(s, s' \mid t)$, while the term $\frac{1}{d_{TD}(s, s' \mid t)^2}$ normalizes each sample loss to be in the range $[0, 1]$. The normalizing term also has the effect of prioritizing pairs of states that are close together on a trajectory, while giving less weight to pairs of states that are further apart. Intuitively, this makes sense since there is more uncertainty regarding the MAD of pairs of states that are further apart on a trajectory.

\subsection{Learning Transition Models}

In the previous section we showed how to learn a state representation that encodes a distance metric between states. This distance allows us to identify states $s_t$ that are close to a given goal state, i.e.~$d(\phi_\theta(s_t), \phi_\theta(s_{goal})) < \epsilon$, or to measure how far we are from the goal state, i.e.~$d(\phi_\theta(s_t), \phi_\theta(s_{goal}))$. However, on its own, the distance metric does not directly give us a policy for reaching the desired goal state.

In this section we propose a method to learn a transition model of actions, that combined with our state representation allows us to plan directly in the embedded space and derive policies to reach any given goal state. Given a dataset of trajectories $\mathcal{T}$ and a state embedding $\phi_\theta(s)$, we seek a parametric transition model $\rho_\zeta(\phi_\theta(s), a)$ such that for any triple $(s, a, s') \in \mathcal{T}$, $\rho_\zeta(\phi_\theta(s), a) \approx \phi_\theta(s')$. 

We propose to learn this model simply by minimizing the squared error as

\begin{equation}
\begin{aligned}
\min_{\zeta} \quad & \sum_t^{\mathcal{T}}\sum_{s, a, s'}^{t} \left[ (\rho_\zeta(\phi_\theta(s), a) - \phi_\theta(s'))^2\right].\\
\end{aligned}
\label{eqn:transition_model}
\end{equation}
Note that in this minimization problem, the parameters $\theta$ of our state representation are fixed, since they are considered known and are thus not optimized at this stage.

\subsection{Latent space planning}

The functions $\rho_\zeta$ and $\phi_\theta$ together represent an approximate model of the underlying MDP.


We propose a Model Predictive Control algorithm that we call Plan-Dist, which computes a policy to reach a given desired goal state $s_{goal} \in S$ by unrolling trajectories for a fixed horizon $H$ in the embedded space. Plan-Dist uses the negative distance between the actual state $s_t$ and the goal state $s_{goal}$ as the desired reward function to be maximized, i.e.~$r(s) = -d(\phi_\theta(s_t), \phi_\theta(s_{goal}))$. Our algorithm considers discrete action spaces and discretizes the action space otherwise. Plan-Dist samples a number $N$ of action trajectories $T_{N, H}$ from the set of all possible action sequences of length $H$, $T_{N,H} \subset A_H$. The trajectories are then unrolled recursively in the latent space starting from our actual state $s_t$ and using the transition model $\phi_\theta(s_{t+1}) \approx \rho_\zeta(\phi_\theta(s_t), a_t)$. At time step $t$, the first action of the trajectory that minimizes the distance to the goal is performed and this process is repeated at each time step until a terminal state is reached (cf.~Algorithm \ref{pc:Plan-Dist}).

\begin{algorithm}[h]
\caption{\textsc{Plan-Dist}} \label{pc:Plan-Dist}
\large
\begin{algorithmic}[1]
	 \STATE {\bf Input}: environment $e$, state embedding $\phi_\theta$, transition model $\rho_\zeta$, horizon $H$, number $N$ of trajectories to evaluate
	 \STATE $s \gets initial state$
	 \STATE $s_{goal} \gets goal state$
	 \STATE $z_{goal} \gets \phi_\theta(s_{goal})$
	 \WHILE{within budget}
	     \STATE $T_{N, H} \gets$ sample $N$ action sequences of length $H$
	     \STATE $t_{MaxReward} \gets None$
	     \STATE $r_{max} \gets MinReward$
	     \FOR{$t_a \in T_{N, H}$}
	     \STATE $z = \phi_\theta(s)$
	     \STATE $r = r - d(z, z_{goal})$
	     \FOR{$a_t \in t_a$}
	     \STATE $z_{t+1} = \rho_\zeta(z, a_t)$
	     \STATE $r = r - d(z_{t+1}, z_{goal})$
	     \ENDFOR
	     \IF{$r > r_{max}$}
	     \STATE $r_{max}=r$
	     \STATE $t_{MaxReward} \gets t_a$
	     \ENDIF
	     \ENDFOR
	     \STATE $s' \gets$ apply action $t_{MaxReward}[0]$ in state $s$
	     \STATE $s = s'$
	\ENDWHILE
 \end{algorithmic}
 \label{algo: manager}
\end{algorithm}

\subsection{Reward Shaping}
Our last contribution is to show how to combine prior knowledge in the form of goal states and our learned distance function to guide existing reinforcement learning algorithms.

We assume that a goal state is given and we augment the environment reward $r(s,a)$ observed by the reinforcement learning agent with Potential-based Reward Shaping~\cite{ng1999policy} of the form:

\begin{equation}
    \overline r(s, a) = r(s, a) + F(s, \gamma, s'),
\end{equation}
where $F$ is our potential-based reward:

\begin{equation*}
    F(s, \gamma, s') = - \gamma d(\phi_\theta(s'), \phi_\theta(s_{goal})) + d(\phi_\theta(s), \phi_\theta(s_{goal})).
\end{equation*}
Here, $d(\phi_\theta(\cdot), \phi_\theta(s_{goal}))$ represents our estimated Minimum Action Distance to the goal $s_{goal}$. Note that for a fixed goal state $s_{goal}$, $-d(\phi_\theta(\cdot), \phi_\theta(s_{goal}))$ is a real-valued function of states which is maximized when $d=0$.

Intuitively our reward shaping schema is forcing the agent to reach the goal state as soon as possible while maximizing the environment reward $r(s, a)$. By using potential-based reward shaping $F(s, \gamma, s')$ we are ensuring that the optimal policy will be invariant~\cite{ng1999policy}.



\section{Experimental Results} \label{Experimental Results}

In this section we present results from experiments where we learn a state embedding and transition model offline from a given dataset of trajectories. We then use the learned models to perform experiments in two settings: 
\begin{enumerate}
    \item Offline goal-conditioned policy learning: Here we evaluate the performance of our Plan-Dist algorithm against GCSL \cite{ghosh2020learning}.
    \item Reward Shaping: In this setting we use the learned MAD distance to reshape the reward of a DDQN\cite{van2016deep} agent (DDQN-PR) for discrete action environments and DDPG\cite{lillicrap2015continuous} for continuos action environment (DDPG-PR), and we compare it to their original versions.
\end{enumerate}
%
Subject to acceptance of the paper, we plan to make the code publicly available to reproduce the experimental results.

\begin{figure}[H]
    \centering
    \includegraphics[width=10cm]{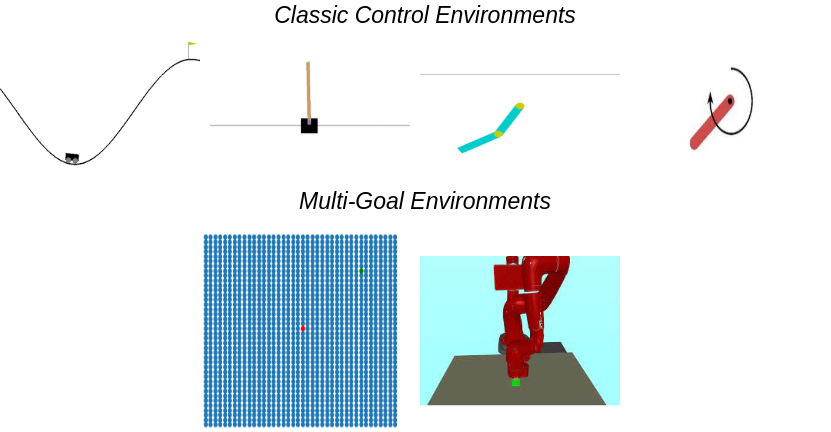}
    \caption{Evaluation Tasks. Top row: MountainCar-v0, CartPole-v0, AcroBot-v1 and Pendulum-v0. Bottom row: GridWorld and SawyerReachXYZEnv-v1.}
    \label{fig:Environments}
\end{figure}

\subsection{Dataset Collection and Domain Description}

We test our algorithms on the classic RL control suite (cf.~Figure \ref{fig:Environments}). Even though termination is often defined for a range of states, we fix a single goal state among the termination states. These domains have complex dynamics and random initial states, making it difficult to reach the goal state without dedicated exploration. The goal state selected for each domain is:

\begin{itemize}
    \item MountainCar-v0: [0.50427865, 0.02712902]
    \item CartPole-v0: [0, 0, 0, 0]
    \item AcroBot-v1: [-0.9661,  0.2581,  0.8875,  0.4607, -1.8354, -5.0000]
    \item Pendulum-v0: [1, 0, 0]
\end{itemize}
Additionally, we test our model-based algorithm Plan-Dist in two multi-goal domains(see. Fig. \ref{fig:Environments}):

\begin{itemize}
    \item A 40x40 GridWorld.
    \item The multiworld domain SawyerReachXYZEnv-v1, where a multi-jointed robotic arm has to reach a given goal position.
\end{itemize}
In each episode, a new goal $s_{goal}$ is sampled at random, so the set of possible goal states $G$ equals the entire state space $S$. These domains are challenging for reinforcement learning algorithms, and even previous work on goal-conditioned reinforcement learning usually considers a small fixed subset of goal states.

In each of these domains we collect a dataset that approximately covers the state space, since we want to be able to use any state as a goal state. Collecting these datasets is not trivial. As an example, consider the MountainCar domain where a car is on a one-dimensional track, positioned between two mountains. A simple random trajectory will not be enough to cover all the state space since it will get stuck in the valley without being able to move the cart on top of the mountains.
Every domain in the classic control suite presents this exploration difficulty and for these environments we rely on collecting trajectories performed by the algorithms DDQN\cite{van2016deep} and DDPG\cite{lillicrap2015continuous} while learning a policy for these domains. Note that we use DDPG only in the Pendulum domain, which is characterized by a continuous action space.

\begin{figure}[H]
    \centering
    \includegraphics[width=12.5cm]{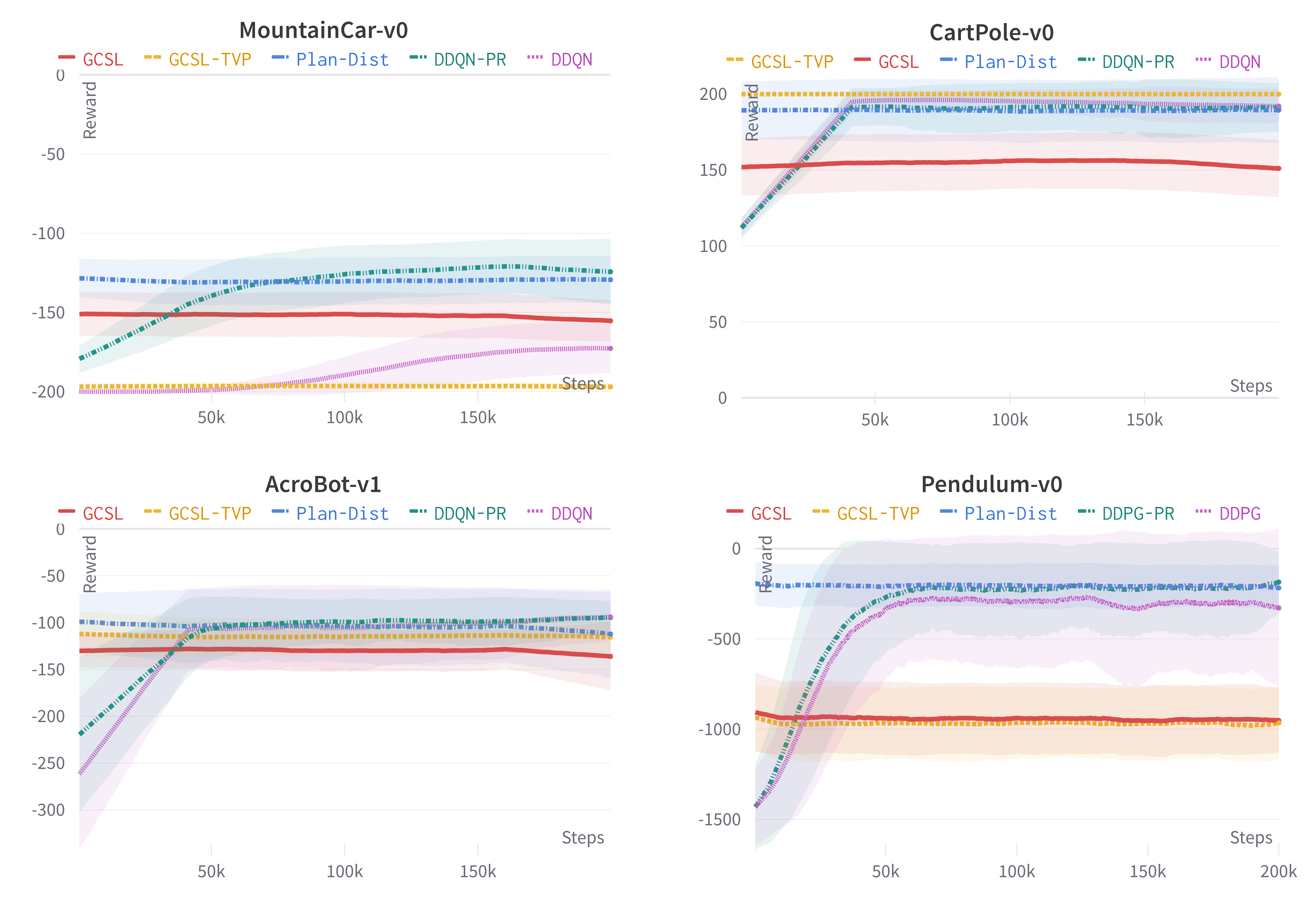}
    \caption{Results in the classic RL control suite.}
    \label{fig:ClassicControl}
\end{figure}

In Table \ref{fig:Datasets} we report the size, the algorithm/policy used to collect the trajectories, the average reward and the maximum reward of each dataset. Note that the average reward is far from optimal and that both Plan-Dist (our offline algorithm) and GCSL improve over the dataset performance (cf.~Figure \ref{fig:ClassicControl}).

\begin{table}[H]
\begin{tabular}{|l|l|l|l|l|}
\hline
{ \textit{\textbf{Environments}}}                             & \textit{\textbf{\begin{tabular}[c]{@{}l@{}}\# Trajectories \\ Dataset\end{tabular}}} & \textit{\textbf{\begin{tabular}[c]{@{}l@{}}Algorithm to \\ Collect Trajectories\end{tabular}}} & \textit{\textbf{\begin{tabular}[c]{@{}l@{}}Avg Reward \\ Dataset\end{tabular}}} & \textit{\textbf{\begin{tabular}[c]{@{}l@{}}Max Reward \\ Dataset\end{tabular}}} \\ \hline
MountainCar-v0 & 100 & DDQN & -164.26 & -112 \\ \hline
CartPole-v0 & 200 & DDQN & +89.42 & +172 \\ \hline
AcroBot-v1 & 100 & DDQN & -158.28 & -92.0 \\ \hline
Pendulum-v0 & 100 & DDPG & -1380.39 & -564.90 \\ \hline
GridWorld & 100 & RandomPolicy & -- & -- \\ \hline
\begin{tabular}[c]{@{}l@{}}SawyerReach-\\ XYZEnv-v1\end{tabular} & 100 & RandomPolicy & -- & -- \\ \hline
\end{tabular}\\
\caption{Dataset description.}%
\label{fig:Datasets}%
\end{table}

\subsection{Learning a State Embedding}

The first step of our procedure consists in learning a state embedding $\phi_\theta$ from a given dataset of trajectories $\mathcal{T}$. From each trajectory $t_i = \{s_0, ..., s_n\} \in \mathcal{T}$ we collect all samples $(s_{i \mid t_i}, s_{j \mid t_i}, d_{TD}(s_{i \mid t_i}, s_{j \mid t_i} \mid t_i))$, $0 \leq i \leq j \leq n$, and populate a Prioritized Experience Replay (PER) memory \cite{schaul2015prioritized}. We use PER to prioritize the samples based on how much they violate our penalty function in \eqref{eqn:relaxed_pf}.


We used mini-batches $B$ of size $512$ with the AdamW optimizer \cite{loshchilov2017decoupled} and a learning rate of $5 * 10^{-4}$ for $100{,}000$ steps to train a neural network $\phi_\theta$ by minimizing the following loss derived from \eqref{eqn:relaxed_pf}:

\begin{equation*}
    \mathcal{L}(\mathcal{B}) = \sum_{(s, s', d_{TD})\in B} \left[ \frac{1}{d_{TD}^2} (\left\|\phi_\theta(s)-\phi_\theta(s')\right\|_{1} - d_{TD})^2\right] + C,
\end{equation*}
where $C$ is the penalty term defined as:

\begin{equation*}
C = \sum_{(s, s', d_{TD})\in B} \left[ \frac{1}{d_{TD}^2} \max(0, \left\|\phi_\theta(s)-\phi_\theta(s')\right\|_{1} - d_{TD})^2\right]
\end{equation*}

We use an embedding dimension of size 64 with an L1 norm as the metric to approximate the MAD distance. Empirically, the L1 norm turns out to perform better than the L2 norm in high-dimensional embedding spaces. These findings are in accordance with theory \cite{aggarwal2001surprising}.

\subsection{Learning Dynamics}

We use the same dataset of trajectories $\mathcal{T}$ to learn a transition model. We collect all the samples $(s, a, s')$ in a dataset $\mathcal{D}$ and train a neural network $\rho_\zeta$ using mini-batches $B$ of size $512$ with the AdamW optimizer \cite{loshchilov2017decoupled} and a learning rate of $5 * 10^{-4}$ for $10{,}000$ steps by mimizing the following loss derived from \eqref{eqn:transition_model}:

\begin{equation*}
    \mathcal{L}(\mathcal{B}) = \sum_{s, s', d_{TD}}^B \left[ (\rho_\zeta(\phi_\theta(s), a) - \phi_\theta(s'))^2 \right]
\end{equation*}

\subsection{Experiments}

We compare our algorithm Plan-Dist against an offline variant of GCSL, where GCSL is trained from the same dataset of trajectories as our models $\phi_\theta$ and $\rho_\zeta$. The GCSL policy and the models $\phi_\theta$ and $\rho_\zeta$ are all learned offline and frozen at test time.

Ghosh et al.~\cite{ghosh2020learning} propose two variants of the GCSL algorithm, a Time-Varying Policy where the policy is conditioned on the remaining horizon $\pi(a | s, g, h)$ (in our experiments we refer to this as GCSL-TVP) and a horizon-less policy $\pi(a | s, g)$ (we refer to this as GCSL).

\begin{figure}[H]
    \centering
    \includegraphics[width=12.5cm]{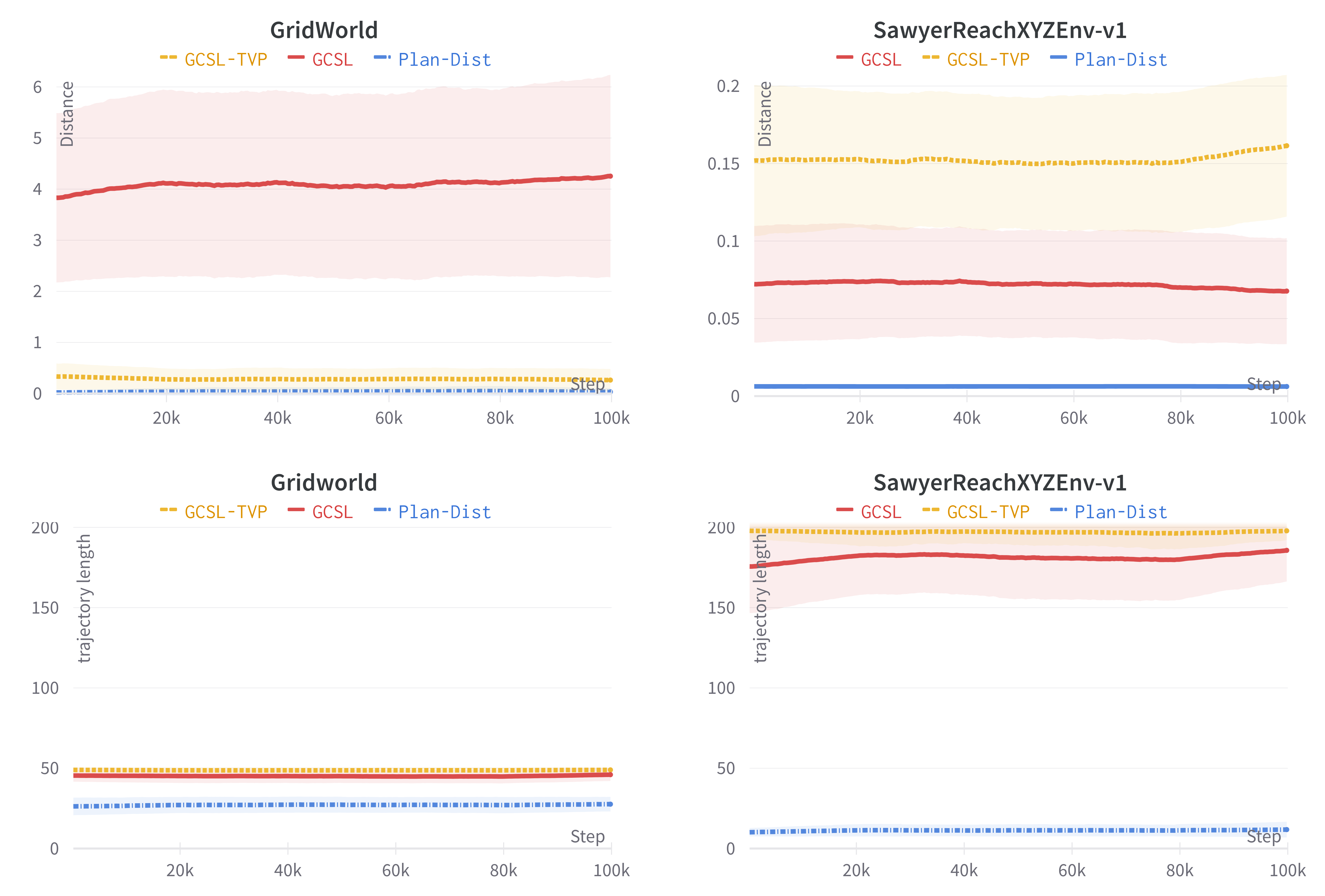}
    \caption{Results in multi-goal environments.}
    \label{fig:MultiGoal}
\end{figure}

We refer to our reward shaping algorithms as DDQN-PR/DDPG-PR and their original counterpart without reward shaping as DDQN/DDPG. DDQN is used in domains in which the action space is discrete, while DDPG is used for continuous action domains.

For all the experiments we report results averaged over 10 seeds where the shaded area represents the standard deviation and the results are smoothed using an average window of length 100. All the hyper-parameters used for each algorithm are reported in the appendix.

In the multi-goal environments in Figure \ref{fig:MultiGoal} we report two metrics: the distance to the goal with respect to the state reached at the end of the episode, and the length of the performed trajectory. In both domains, the episode terminates either when we reach the goal state or when we reach the maximum number of steps ($50$ steps for GridWorld, and $200$ steps for SawyerReachXYZEnv-v1). We evaluate the algorithms for $100{,}000$ environment steps. 

We can observe that Plan-Dist is able to outperform GCSL, being able to reach the desired goal state with better precision and by using shorter paths. We do not compare to reinforcement learning algorithms in these domains since they struggle to generalize when the goal changes so frequently.

On the classic RL control suite in Figure \ref{fig:ClassicControl} we report the results showing the total reward achieved at the end of each episode. Here we compare both goal-conditioned algorithms and state-of-the-art reinforcement learning algorithms for $200{,}000$ environment steps. Plan-Dist is still able to outperform GCSL in almost all domains, while performing slightly worse than GCSL-TVP in CartPole-v0. Compared to DDQN-PR/DDPG-PR, Plan-Dist is able to reach similar total reward, but in MountainCar-v0, DDQN-PR is eventually able to achieve higher reward.

The reward shaping mechanism of DDQN-PR/DDPG-PR is not helping in the domains CartPole-v0, Pendulum-v0 and Acrobot-v0. In these domains, it is hard to define a single state as the goal to reach in each episode. As an example, in CartPole-v0 we defined the state $[0, 0, 0, 0]$ as our goal state and we reshape the reward accordingly, but this is not in line with the environment reward that instead cares only about balancing the pole regardless of the position of the cart. While in these domains we do not observe an improvement in performance, it is worth noticing that our reward shaping scheme is not adversely affecting DDQN-PR/DDPG-PR, and they are able to achieve results that are similar to those of their original counterparts.

Conversely, in MountainCar-v0 where the environment reward resembles a goal reaching objective, since the goal is to reach the peak of the mountain as fast as possible, our reward shaping scheme is aligned with the environment objective and DDQN-PR outperforms DDQN in terms of learning speed and total reward on the fixed evaluation time of $200{,}000$ steps.

\section{Discussion and Future Work}

We propose a novel method for learning a parametric state embedding $\phi_\theta$ where the distance between any pair of states $(s, s')$ in embedded space approximates the Minimum Action Distance, $d(\phi_\theta(s), \phi_\theta(s')) \approx d_{MAD}(s, s')$. One limitation of our approach is that we consider symmetric distance functions, while in general the MAD in an MDP could be asymmetric, $d_{MAD}(s, s') \neq d_{MAD}(s', s)$. Schaul et al.~\cite{schaul2015universal} raise a similar issue in the context of learning Universal Value Functions, and propose an asymmetric distance function on the following form:

\begin{equation*}
    d_A(s, s') = \| \sigma(\psi_1(s')) (\phi(s) - \psi_2(s')) \|_l,
\end{equation*}
where $\sigma$ is a the logistic function and $\psi_1$ and $\psi_2$ are two halves of the same embedding vector. In their work they show similar performance using the symmetric and asymmetric distance functions. Still, an interesting future direction would be to use this asymmetric distance function in the context of our self-supervised training scheme.


While our work focuses on estimating the MAD between states and empirically shows the utility of the resulting metric for goal-conditioned reinforcement learning, the distance measure could be uninformative in highly stochastic environment where the expected shortest path distance better measures the distance between states. One possible way to approximate this measure using our self-supervised training scheme would be to minimize a weighted version of our objective in \eqref{eqn:pf}:

\begin{equation*}
\min_{\theta} \quad \sum_{t\in\mathcal{T}}\sum_{(s,s')\in t}  1/d_{TD}^\alpha (\left\|\phi_\theta(s)-\phi_\theta(s')\right\|_{l} - d_{TD}(s, s' \mid t))^2.
\end{equation*}
Here, the term $1/d_{TD}$ is exponentiated by a factor $\alpha$ which decides whether to favour the regression over shorter or longer Trajectory Distances. Concretely, when $\alpha<1$ we favour the regression over shorter Trajectory Distances, approximating a Shortest Path Distance.

In our work we learn a distance function offline from a given dataset of trajectories, and one possible line of future research would be to collect trajectories while simultaneously exploring the environment in order to learn the distance function. 

In this work we focus on single goal reaching tasks, in order to have a fair comparison with goal-conditioned reinforcement learning agents in the literature. However, the use of our learned distance function is not limited to this setting and we can consider multi-goal tasks, such as reaching a goal while maximizing the distance to forbidden (obstacle) states, reaching the nearest of two goals, and in general any linear and non-linear combination of distances to states given as input.

Lastly, it would be interesting to use this work in the contest of Hierarchical Reinforcement Learning, in which a manager could suggest subgoals to our Plan-Dist algorithm.

\bibliographystyle{splncs04}
\bibliography{bibliography/reference.bib}
\newpage
\appendix

\section{Hyperparameters}

In Table \ref{tab:1} we report all hyperparameters used to train our models $\phi_\theta$, $\rho_\zeta$ and hyperparameters used for Plan-Dist.

\begin{table}[H]
\begin{tabular}{|ll|ll|ll|}
\hline
\multicolumn{2}{|l|}{{\ul \textit{\textbf{Compression Function $\phi_\theta$}}}} & \multicolumn{2}{l|}{{\ul \textit{\textbf{Transition Model $\rho_\zeta$}}}} & \multicolumn{2}{l|}{{\ul \textit{\textbf{Plan-Dist Algorithm}}}} \\ \hline
\multicolumn{1}{|l|}{Hyperparameters} & Value & \multicolumn{1}{l|}{Hyperparameters} & Value & \multicolumn{1}{l|}{Hyperparameters} & Value \\ \hline
\multicolumn{1}{|l|}{Neural Network} & \begin{tabular}[c]{@{}l@{}}FC1(128)\\ FC2(128)\end{tabular} & \multicolumn{1}{l|}{Neural Network $z$} & \begin{tabular}[c]{@{}l@{}}FC1(128)\\ FC2(128)\end{tabular} & \multicolumn{1}{l|}{H} & 5 \\ \hline
\multicolumn{1}{|l|}{Activation} & Selu & \multicolumn{1}{l|}{Neural Network $a$} & \begin{tabular}[c]{@{}l@{}}FC1(128)\\ FC1(128)\end{tabular} & \multicolumn{1}{l|}{N random traj} & 20 \\ \hline
\multicolumn{1}{|l|}{Embedding Dimension} & 64 & \multicolumn{1}{l|}{Neural Network $z + a$} & \begin{tabular}[c]{@{}l@{}}FC1(128)\\ FC1(128)\end{tabular} & \multicolumn{1}{l|}{} &  \\ \hline
\multicolumn{1}{|l|}{Trainin steps} & 100000 & \multicolumn{1}{l|}{Activation} & Selu & \multicolumn{1}{l|}{} &  \\ \hline
\multicolumn{1}{|l|}{Batch size} & 512 & \multicolumn{1}{l|}{Embedding Dimension} & 64 & \multicolumn{1}{l|}{} &  \\ \hline
\multicolumn{1}{|l|}{Learning Rate} & $5*10^-4$ & \multicolumn{1}{l|}{Trainin steps} & 10000 & \multicolumn{1}{l|}{} &  \\ \hline
\multicolumn{1}{|l|}{Optimizer} & AdamW & \multicolumn{1}{l|}{Batch size} & 512 & \multicolumn{1}{l|}{} &  \\ \hline
\multicolumn{1}{|l|}{PER $\epsilon$} & 0.1 & \multicolumn{1}{l|}{Learning Rate} & $5*10^-4$ & \multicolumn{1}{l|}{} &  \\ \hline
\multicolumn{1}{|l|}{PER $\alpha$} & 0.6 & \multicolumn{1}{l|}{Optimizer} & AdamW & \multicolumn{1}{l|}{} &  \\ \hline
\end{tabular}
\caption{Hyperparameters for our models $\phi_\theta$, $\rho_\zeta$ and our algorithm Plan-Dist.}
\label{tab:1}
\end{table}

\newpage
In Table \ref{tab:2} we report all hyperparameters used to train our baselines.

\begin{table}[H]
\begin{tabular}{|ll|ll|ll|}
\hline
\multicolumn{2}{|l|}{{\ul \textit{\textbf{DDQN}}}} & \multicolumn{2}{l|}{{\ul \textit{\textbf{DDPG}}}} & \multicolumn{2}{l|}{{\ul \textit{\textbf{GCSL}}}} \\ \hline
\multicolumn{1}{|l|}{Hyperparameters} & Value & \multicolumn{1}{l|}{Hyperparameters} & Value & \multicolumn{1}{l|}{Hyperparameters} & Value \\ \hline
\multicolumn{1}{|l|}{Neural Network} & \begin{tabular}[c]{@{}l@{}}FC1(32)\\ FC2(32)\\ FC3(32)\end{tabular} & \multicolumn{1}{l|}{\begin{tabular}[c]{@{}l@{}}Neural Network \\ Actor and Critic\\ (separated)\end{tabular}} & \begin{tabular}[c]{@{}l@{}}FC1(32)\\ FC2(32)\\ FC3(32)\end{tabular} & \multicolumn{1}{l|}{Neural Network} & \begin{tabular}[c]{@{}l@{}}FC1(128)\\ FC2(128)\end{tabular} \\ \hline
\multicolumn{1}{|l|}{Activation} & Selu & \multicolumn{1}{l|}{Activation} & Selu & \multicolumn{1}{l|}{Activation} & Selu \\ \hline
\multicolumn{1}{|l|}{Optimizer} & Adam & \multicolumn{1}{l|}{Optimizer} & Adam & \multicolumn{1}{l|}{Trainin steps} & 100000 \\ \hline
\multicolumn{1}{|l|}{Trainin steps} & 100000 & \multicolumn{1}{l|}{Activation} & Selu & \multicolumn{1}{l|}{Batch size} & 256 \\ \hline
\multicolumn{1}{|l|}{discount factor} & 0.99 & \multicolumn{1}{l|}{discount factor} & 0.99 & \multicolumn{1}{l|}{Learning Rate} & $5*10^{-4}$ \\ \hline
\multicolumn{1}{|l|}{Learning Rate} & 0.001 & \multicolumn{1}{l|}{Learning Rate} & 0.001 & \multicolumn{1}{l|}{Optimizer} & Adam \\ \hline
\multicolumn{1}{|l|}{TargetNetworkUpdate} & Poliak 0.1 & \multicolumn{1}{l|}{TargetNetworkUpdate} & Poliak 0.1 & \multicolumn{1}{l|}{} &  \\ \hline
\multicolumn{1}{|l|}{epsilon\_decay (at each step)} & 0.999 & \multicolumn{1}{l|}{Batch Size} & 100 & \multicolumn{1}{l|}{} &  \\ \hline
\multicolumn{1}{|l|}{Batch Size} & 100 & \multicolumn{1}{l|}{} &  & \multicolumn{1}{l|}{} &  \\ \hline
\end{tabular}
\caption{Hyperparameters for our baselines DDQN, DDPG, GCSL.}
\label{tab:2}
\end{table}

\end{document}